\documentclass[conference]{IEEEtran}

\usepackage{tabularx,booktabs}
\usepackage{etoolbox}
\makeatletter
\patchcmd{\@makecaption}
  {\scshape}
  {}
  {}
  {}
\makeatother
\usepackage{multirow}

\usepackage{ifpdf}
 \ifpdf
 \else
 \fi

\usepackage[noadjust]{cite}

\ifCLASSINFOpdf
   \usepackage[pdftex]{graphicx}
\else
\fi

\usepackage{amsmath}
\usepackage{amssymb}

\usepackage{algorithmic}
\usepackage{algorithm}

\usepackage{mfirstuc} 
\usepackage{titlesec}

\usepackage{array}

\ifCLASSOPTIONcompsoc 
  \usepackage[caption=false,font=normalsize,labelfont=sf,textfont=sf]{subfig}
\else
  \usepackage[caption=false,font=footnotesize]{subfig}
\fi

\usepackage{url}

\makeatletter
\let\NAT@parse\undefined
\makeatother
\usepackage{hyperref}  

\usepackage{textcomp}
\usepackage{xcolor}
\usepackage{color}
\usepackage{fancyhdr}
\def\BibTeX{{\rm B\kern-.05em{\sc i\kern-.025em b}\kern-.08em
    T\kern-.1667em\lower.7ex\hbox{E}\kern-.125emX}}
\begin{document}

\title{
LineGS : 3D Line Segment Representation on 3D Gaussian Splatting
}

\author{\IEEEauthorblockN{Yang Chenggang}
\IEEEauthorblockA{\textit{School of Computing} \\
\textit{National University of Singapore}\\
\href{mailto:cyang_09@u.nus.edu}{cyang\_09@u.nus.edu}}
\and
\IEEEauthorblockN{Shi Yuang}
\IEEEauthorblockA{\textit{School of Computing} \\
\textit{National University of Singapore}\\
\href{mailto:yuangshi@u.nus.edu}{yuangshi@u.nus.edu}}
}

\maketitle
\thispagestyle{fancy} 
\lhead{} 
\chead{} 
\rhead{} 
\lfoot{} 
\cfoot{} 
\rfoot{} 
\renewcommand{\headrulewidth}{0pt} 
\renewcommand{\footrulewidth}{0pt} 
\pagestyle{fancy}
\cfoot{\thepage}
\begin{abstract}
Abstract representations of 3D scenes play a crucial role in computer vision, enabling a wide range of applications such as mapping, localization, surface reconstruction, and even advanced tasks like SLAM and rendering. Among these representations, line segments are widely used because of their ability to succinctly capture the structural features of a scene. However, existing 3D reconstruction methods often face significant challenges. Methods relying on 2D projections suffer from instability caused by errors in multi-view matching and occlusions, while direct 3D approaches are hampered by noise and sparsity in 3D point cloud data.
This paper introduces LineGS, a novel method that combines geometry-guided 3D line reconstruction with a 3D Gaussian splatting model to address these challenges and improve representation ability. The method leverages the high-density Gaussian point distributions along the edge of the scene to refine and optimize initial line segments generated from traditional geometric approaches. By aligning these segments with the underlying geometric features of the scene, LineGS achieves a more precise and reliable representation of 3D structures. The results show significant improvements in both geometric accuracy and model compactness compared to baseline methods. The code is released at \textit{\textcolor[rgb]{1, 0.5, 0.75}{\href{https://github.com/ericshenggle/LineGS}{https://github.com/ericshenggle/LineGS}}}.
\end{abstract}
\section{Introduction}

Identifying spatial geometric information that describes a 3D scene and constructing an abstract representation from it holds great significance in 3D computer vision. Such representations can support various critical tasks. Common primitives like line segments and curves capture edge information within scenes and are used for mapping~\cite{bartoli2005structure, liu20233d, ramalingam2015line}, localization~\cite{hruby2024handbook, lee2019elaborate, liu1990determination}, surface reconstruction enhancement~\cite{huang2013edge}, Simultaneous Localization and Mapping (SLAM)~\cite{smith2006real}, and rendering~\cite{celes2010texture, celes2011fast}. Other primitives, such as point clouds and patches, describe the 3D structure of a scene and are applied in tasks like reconstruction, depth estimation~\cite{cheng2018depth,cui2021deep} and completion~\cite{ma2019self, hu2021penet}.

3D line segment reconstruction offers a straightforward form of abstract representation, using multiple line segments of varying lengths, angles, and positions to depict a 3D scene. Many seminal methods first identify matching feature points in 2D images~\cite{heinly2015reconstructing, schoenberger2016sfm} and convert them into 2D line segments~\cite{ipol.2012.gjmr-lsd, huang2020tp, zhang2021elsd}, which are then projected back into 3D space using geometric algorithms. This approach captures prominent line information in the scene, typically representing spatial or color boundaries of objects, and these recent geometry-guided methods~\cite{hofer2017efficient, pautrat2021sold2, wei2022elsr, pautrat2023deeplsd, liu20233d}, are known for their high performance and effectiveness. However, processing 2D information inherently lacks repeatability, which can result in stability issues and reduce the robustness of the final 3D outcome. 

3D line reconstruction can also be achieved directly from 3D feature information, bypassing limitations imposed by geometric projection methods. For instance, dense point cloud features obtained from Sructure-From-Motion (SFM)~\cite{schoenberger2016sfm} can be used to extract 3D line segments. Typically, this involves identifying or classifying 3D points near boundaries, then performing clustering analysis on their spatial distribution to abstract the line segments. However, in current methods, boundary points make up only a small portion of the initial point cloud, with the remaining points adding significant noise, making the process challenging. Some trained frameworks~\cite{wang2020pie, liu2021pc2wf} can achieve good fits within specific scene types, while other methods~\cite{ye2023nef, xue2024neat} first extract point clouds near boundaries before further processing.

To combine the strengths of both approaches, a straightforward method is to use the result of one to refine and optimize the other. In this paper, we proposes a simple method called LineGS that integrates geometry-based line reconstruction with 3D Gaussian splatting model~\cite{kerbl3Dgaussians} to achieve more accurate 3D line reconstruction that better represents 3D feature information. The 3D Gaussian splatting model consists of numerous points with Gaussian parameters, carrying rich geometric information, and is known for its high performance, ease of training, and effectiveness. While the Gaussian model offers a robust representation of 3D feature information in the scene, geometry-based line reconstruction can also cover these features. Thus, using the 3D feature information embedded in a trained Gaussian model, we can effectively post-process the initial 3D line segments from geometry-based methods, ultimately providing an abstracted line-based representation of Gaussian points within the Gaussian model. Gaussian points tend to cluster along geometric and color boundaries. By adjusting the line segments based on Gaussian density, we retain the overall geometric structure of the segments while aligning them more closely with the distribution of Gaussian points in the model.

In summary, our work makes the following contributions: i) We propose a simple and efficient 3D line segment processing method that combines the efficiency of geometric approaches with the robustness of 3D feature points. ii) Leveraging the spatial information of Gaussian points, we introduce Gaussian density-based adjustments to refine the position, direction, and length of line segments, enhancing the accuracy and simplicity of the final result. iii) We provide an approach to evaluate the abstract representation ability of 3D line segments with respect to the 3D Gaussian model, validating the effectiveness of our method based on this evaluation.
\section{Related Work}

Early 3D line reconstruction methods relied heavily on Structure-from-Motion (SfM) frameworks to detect and match lines across multiple images, where line detection was achieved through descriptors such as LSD~\cite{ipol.2012.gjmr-lsd} and later improved by~\cite{huang2020tp, pautrat2023deeplsd}. These descriptors are integral to matching lines across views, which remain challenging due to occlusions and varying line endpoints. Works like Bartoli et al.~\cite{bartoli2004framework,bartoli2005structure} initiated full SfM pipelines that integrate line segment detection with matching mechanisms, while Schindler~\cite{schindler2006line} introduced the Manhattan-world assumption to add geometric constraints, enhancing robustness in structured environments. For lifting matched lines into 3D, traditional methods often employ triangulation techniques~\cite{baillard1999automatic, chandraker2009moving, micusik2017structure, ramalingam2015line, zhang2014structure}. However, triangulation alone can struggle with scenes containing repetitive or complex structures. Epipolar geometry has also been leveraged to reconstruct lines with greater geometric consistency, as seen in works like Hofer et al. ~\cite{hofer2014improving, hofer2015line3d, hofer2017efficient}, who introduced weak epipolar constraints to refine line matches, later integrated into the Line3D++ framework, which remains a robust choice for obtaining high-quality 3D line maps. Additionally, methods like ELSR~\cite{wei2022elsr, liu20233d} have made impressive progress in addressing the limitations of geometry-based approaches, the line detector robustness and matching performance remains a bottleneck. Depth-based approaches offer an alternative to multi-view matching by directly detecting and projecting lines from depth maps. Zhang et al.~\cite{fang20233} and others~\cite{kahl2003multiview, robert1991curve, schmid2000geometry} demonstrated depth-driven techniques where 2D line segments detected in depth images are directly transformed into 3D line segments. These methods simplify the reconstruction process but face challenges when noise is present in depth data. 

In addition to geometry-based methods, several approaches directly operate on 3D point clouds to reconstruct line structures, bypassing the challenges of multi-view image matching. These methods typically start by classifying edge points within the point cloud, clustering points that belong to the same edge, and fitting parametric lines to these clusters. However, these methods are sensitive to noise and require careful parameter tuning. To further mitigate noise and imbalanced edge point distribution in 3D point clouds, several works~\cite{chen2017fast, schnabel2007efficient, wang2020pie, ye2023nef} have introduced point filtering, weighted loss functions, and robust edge parameter estimation, improving line classification accuracy and robustness. Recent advancements have seen a shift toward end-to-end learning-based frameworks that predict 3D line structures directly from images, bypassing the need for explicit matching. Methods~\cite{jain2010exploiting, ma20223d} predict 3D wireframes, a set of interconnected 3D lines forming a graph with vertices at junctions, and learning-based techniques~\cite{zhou2019learning, liu2021pc2wf, luo2022learning, pautrat2021sold2} have also been explored to refine these wireframe predictions. Chelani et al.~\cite{chelani2024edgegaussians} use 2D edge map to construct Gaussians model that only contains the points that on the edges and then cluster the points to parametric edges. These methods provide high-quality line reconstructions in structured scenes but often require large, annotated datasets to achieve generalization.

While geometric methods provide a foundation for 3D line reconstruction, each approach—whether line descriptors, depth-based, or learning models—has unique strengths and limitations. Point-based methods, for instance, struggle with noise, making it hard to distinguish edges through clustering. Instead, our approach combines geometry-based line reconstruction with Gaussian points: we use initial segments from geometric methods as edges, then refine them based on Gaussian point distribution. This yields an abstract representation that more accurately reflects the spatial distribution of the Gaussian model.
\section{3D Line Representation with Gaussians}
\label{sec:3.0}
The 3D Gaussian splatting model is highly efficient and capable in both rendering and training processes~\cite{kerbl3Dgaussians}. And centers of these 3D Gaussian points are concentrated along the edges of objects and pixels within the actual 3D scene. These edges can be first efficiently derived by using geometry-guided line segment reconstruction methods~\cite{hofer2017efficient, wei2022elsr}. As a result, the derived line segments can be refined and optimized using the trained Gaussian splatting model.

In this section, we first describe how the 3D Gaussian model represents a scene and its robust spatial characteristics in 3D space. We then discuss the limitations of existing line reconstruction methods in accurately capturing the scene structure and representing the 3D Gaussian model. Finally, we present an approach to obtain more precise 3D line representations for 3D Gaussians by combining the initial line segments from existing reconstruction methods with trained 3D Gaussian points.

\subsection{Preliminaries: 3D Gaussian Splatting}
\label{sec:3.1}

3D Gaussian Splatting is a alternative representation that can render 3D scene from arbitrary view using a set of 3D Gaussians. Each Gaussian contains the 3D coordinate information and higher-order spherical harmonic coefficients which defines its opacity and color. A Gaussian centered at 3d point $\mu \in \mathbb{R}^{3\times3}$ is defined by a full covariance matrix $\Sigma$ :
\begin{equation}
\label{equ:gaussian defination}
        G(x) = e^{-\frac{1}{2}(x)^T\Sigma^{-1}(x)}
\end{equation}
And rendering part need to project these 3D Gaussians to 2D image space. Previous work \cite{zwicker2001ewa} show that to do this projection, the covariance matrix $\Sigma$ in camera coordinates can be defined by a viewing transformation and the Jacobian of the affine approximation of the projective transformation. The parameters of the Gaussian model are differentiable, and the 2D image rendered through training will be compared and loss calculated with the original image. To optimize the $\Sigma$ while maintain its positive semi-definiteness that can only represent its physical meaning, the covariance matrix $\Sigma$ is defined as $\Sigma = RSS^TR^T$, where $R \in \mathbb{R}^{3\times3}$ is a rotation matrix and $S \in \mathbb{R}^{3\times3}$ is a scaling matrix.

The initialization of the model is a set of Gaussians centered at a sparse point cloud that obtained SfM~\cite{schoenberger2016sfm}. These Gaussian points will be split, duplicated, or pruned based on the difference between the splashing shape rendered as a 2D ellipse and the actual image. In other words, the points will be denser in areas that require a higher level of fitting and sparser in areas where a lower level of fitting is needed. And areas of higher level of fitting are typically located at the edges of certain scenes, the protruding parts of 3D objects, or pixel boundaries with significant color differences. For instance, they might appear along the edge of a table or the black gaps between strips on a brown floor. 
\begin{figure}[h]	
    \centering	
    \subfloat[Original image]{\includegraphics[width=0.49\linewidth]{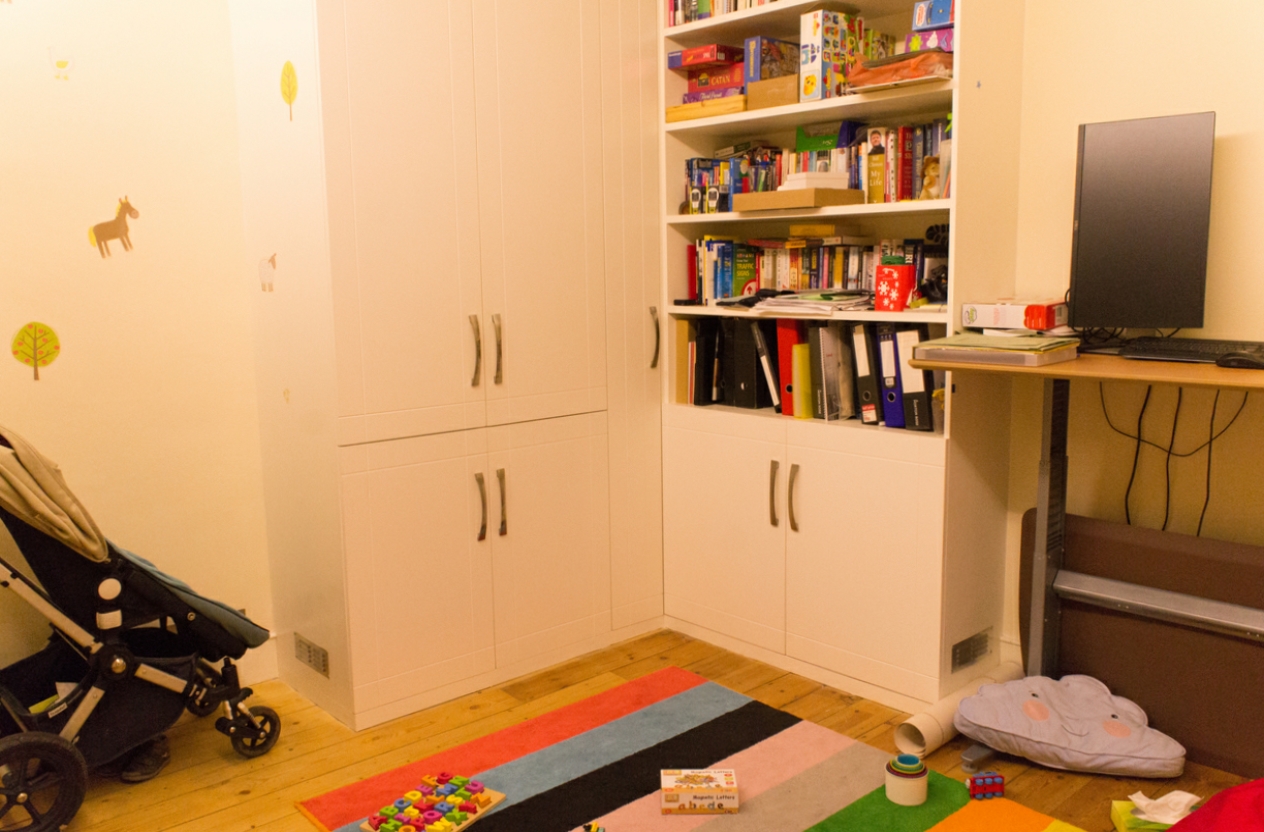}}
    \subfloat[Gaussian points]{\includegraphics[width=0.49\linewidth]{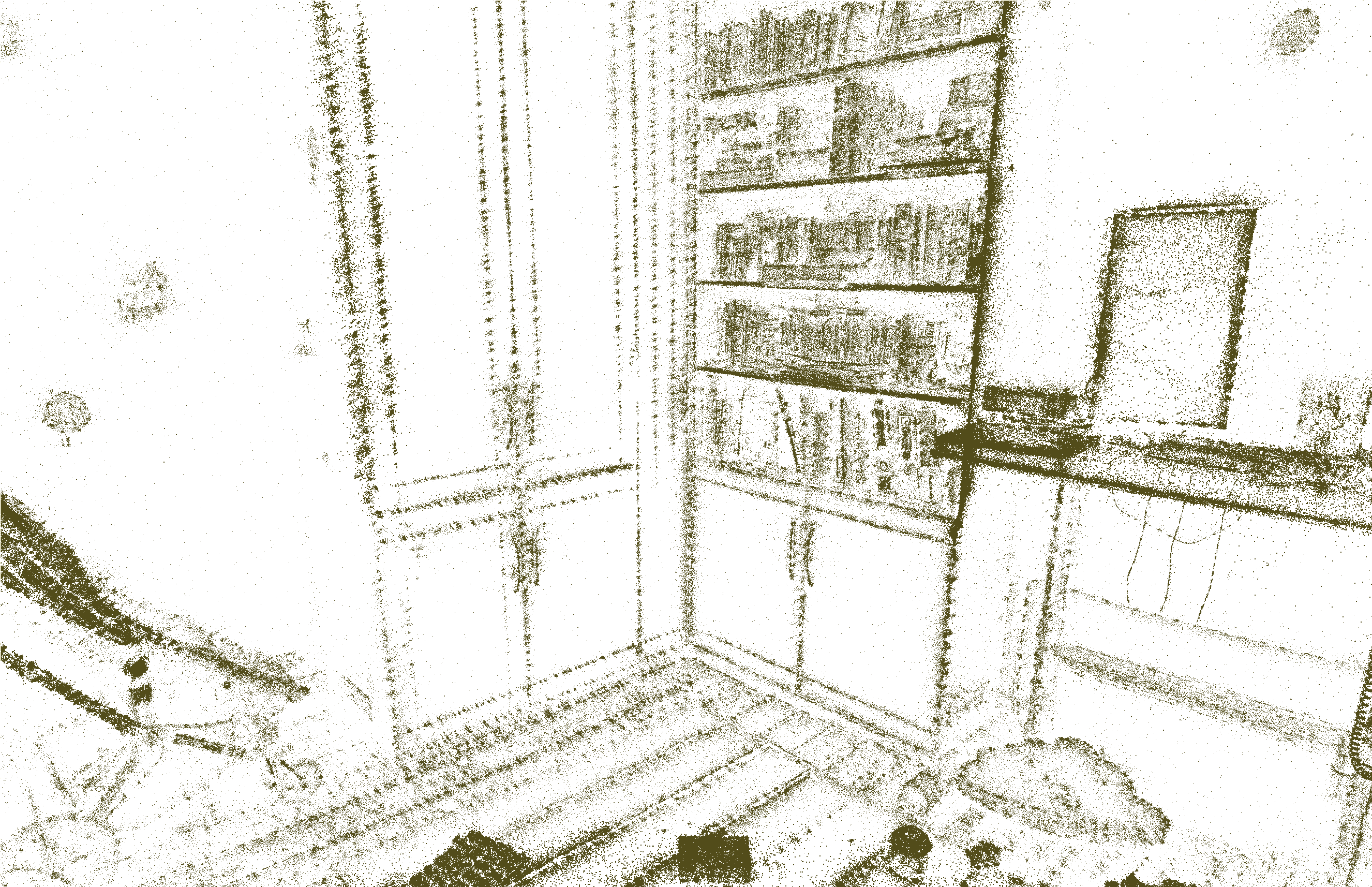}}	\\
    \caption{\textbf{Distribution of trained Gaussian points.} The centers of Gaussian points are concentrated at the boundaries of pixel colors and preserve intact three-dimensional spatial information. This scene is $\mathcal{P}$LAYROOM from the Deep Blending dataset~\cite{hedman2018deep}.}	
    \label{fig:playroom}
\end{figure}

In Fig.\ref{fig:playroom}, the scene has a white cabinet and various colorful books placed on it, as well as tables and colorful carpets. It can be clearly seen that at the boundaries of these colorful objects, the centers of Gaussian points are concentrated in these areas. And in the very similar pixel area of a large block, $e.g.$, the flat surface of the white wardrobe, there are only very sparse Gaussians. With these characteristics, we know that Gaussian points distributed along spatial and color pixel boundaries share similar gaussian features, as they all adapt closely to the boundary and maintain spatial consistency, with Gaussian points clustering along various parts of the boundary. Finding a good representation of these Gaussians can help some further work on Gaussian splatting model.

\subsection{Geometry-guided Line Segment Reconstruction Issues}
\label{sec:3.2}

These geometry-guided 3D line reconstruction methods~\cite{hofer2017efficient, wei2022elsr} leverage advanced geometric computation algorithms to efficiently reconstruct 3D line segments from a set of images and sparse point cloud inputs, which also serve as the input for 3D Gaussians. These reconstruction methods are both fast and effective, producing high-quality results. Typically, they use two images from different viewpoints to process and match~\cite{ipol.2012.gjmr-lsd}, obtaining reliable 2D line segments~\cite{felzenszwalb2004efficient}. These segments are then back-projected into 3D space using various geometric algorithms. By utilizing multiple sets of images from different viewpoints, numerous 3D line segments can be reconstructed, classified, and merged~\cite{donoser2013replicator}. However, due to the limitations of this unsupervised approach and inherent algorithmic errors, the resulting line segments often lack robustness and reproducibility.
\begin{figure}[htbp]	
    \centering	
    \subfloat[Original]{\includegraphics[width=0.49\linewidth]{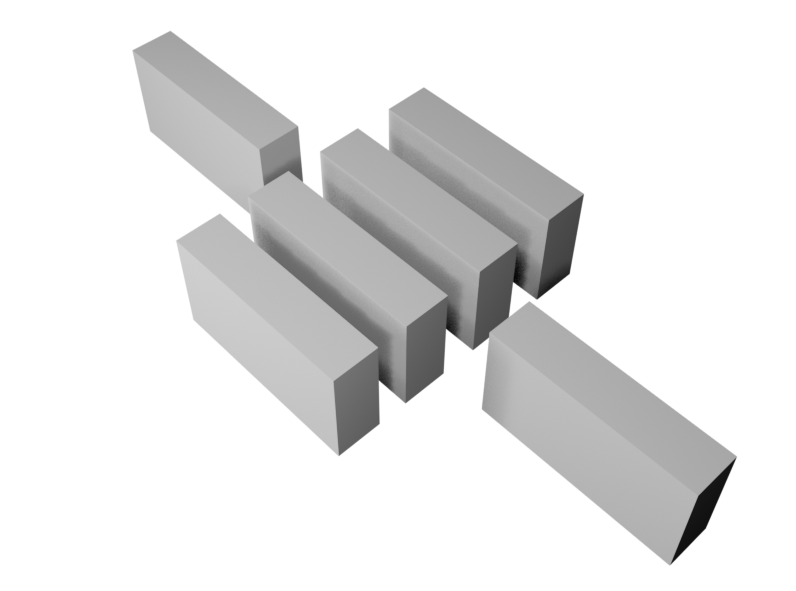}}
    \subfloat[Gaussian points]{\includegraphics[width=0.49\linewidth]{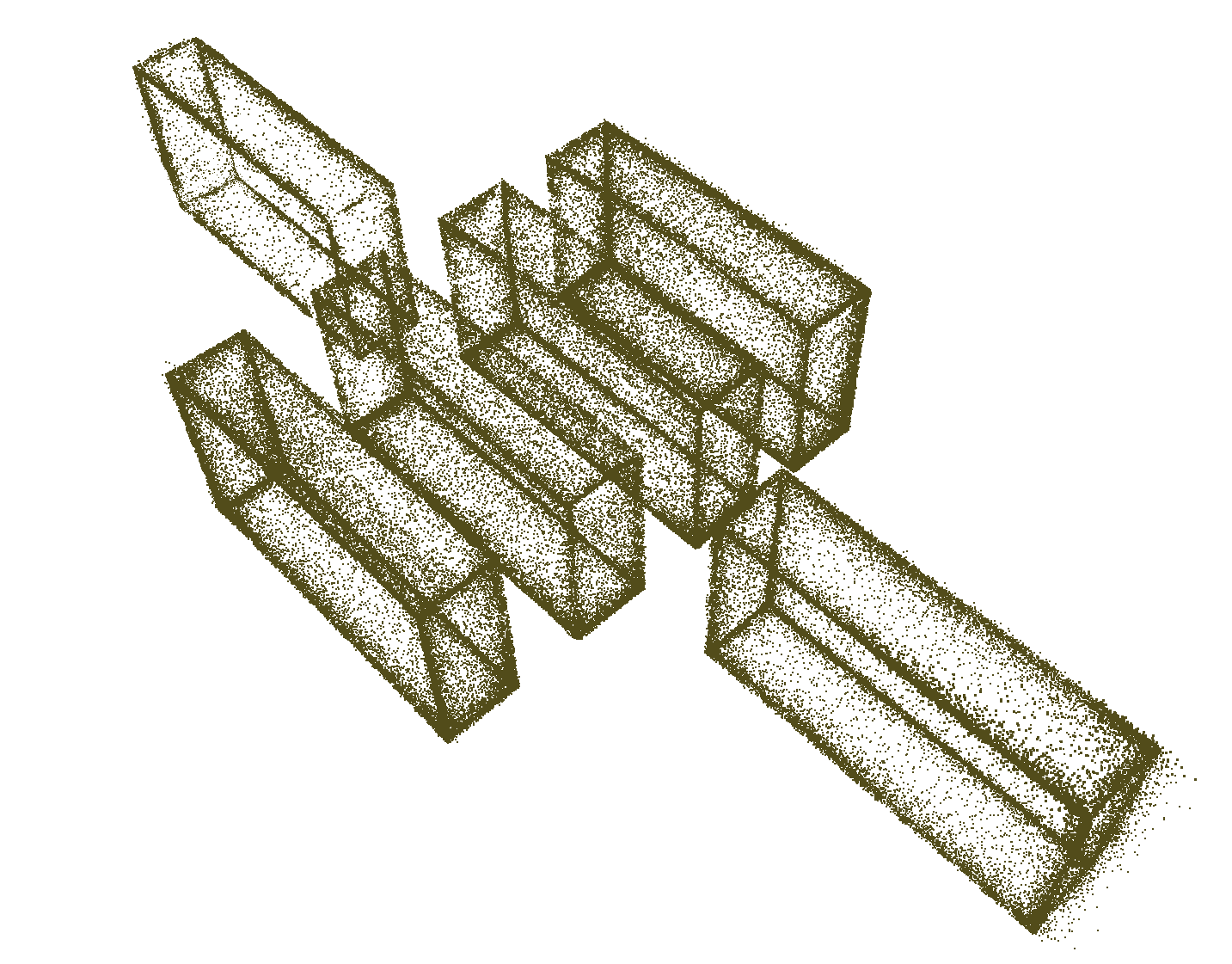}}	\\
    \subfloat[Issues]{\includegraphics[width=\linewidth]{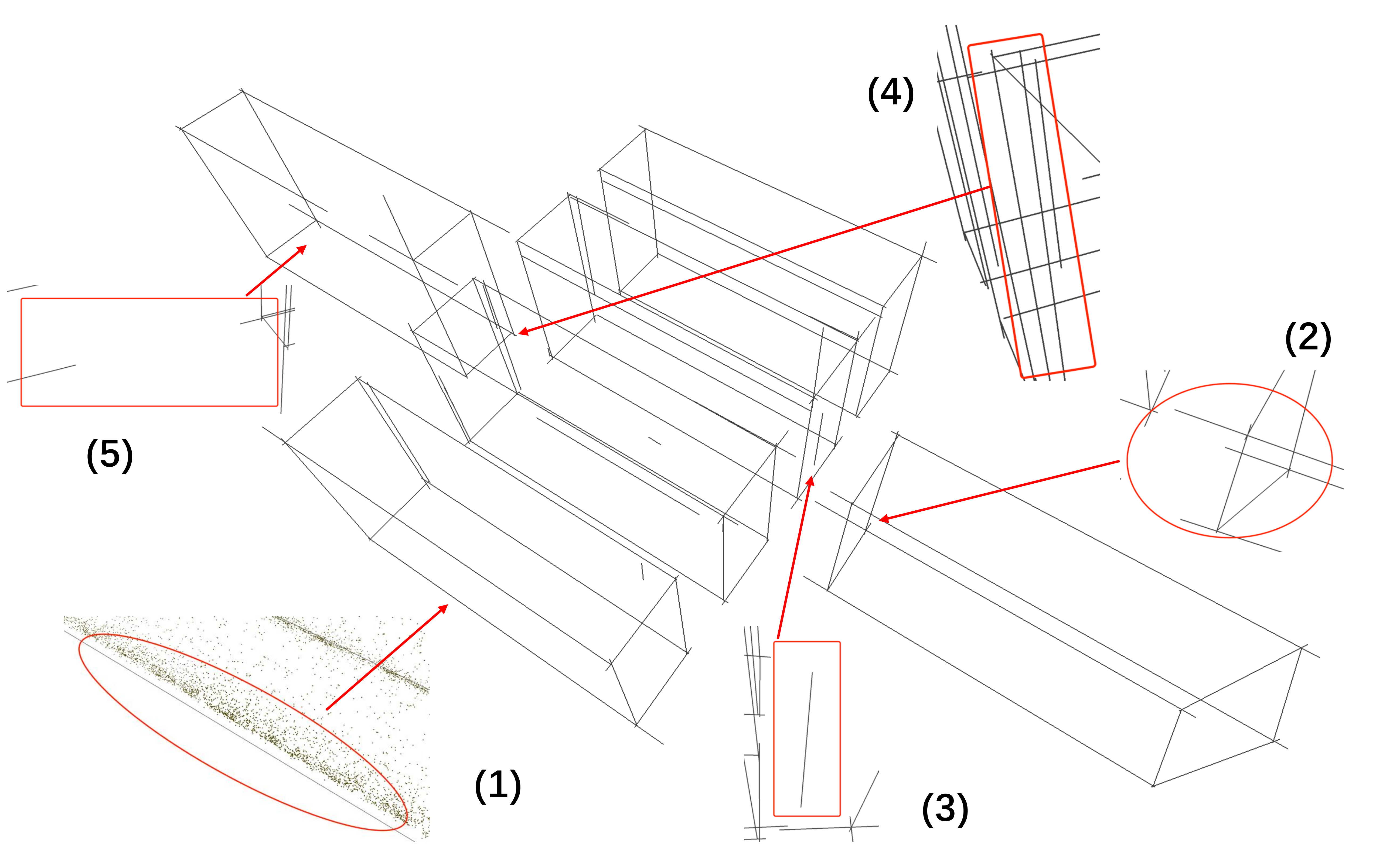}}
    \caption{\textbf{The visualization of these issues}. Given a trick example, constructed using Blender, arranges multiple quadrilaterals together and renders 100 images around the center of the scene}	
    \label{fig:line_example}
\end{figure}

Specifically, when combined with Gaussians, the generated line segments typically exhibit the following issues, and there is an visual example in Fig.~\ref{fig:line_example}.
\begin{enumerate}
    \item \textbf{Position Bias.} The 3DGS centers are not precisely located on sharp areas due to the characteristics of their splatting model.
    \item \textbf{Overextension.} Errors arise when a 3D segment is positioned correctly but has an incorrect length, calculated from a pair of matched 2D images.
    \item \textbf{Outliers.} Errors occur when a 3D segment has an incorrect position, derived from a pair of matched 2D images.
    \item \textbf{Duplication.} Multiple similar segments may not be properly clustered and merged when generated from different pairs of matched 2D images, resulting in unrepeatable and non-robust outputs.
    \item \textbf{Discontinuity.} Discontinuities may occur in segments generated from a pair of matched 2D images, particularly in areas with curves.
\end{enumerate}
Depending on the quantity of input images and the varying camera viewpoints, the generated line segments will have different error proportions. However, these errors are typically trade-offs, and it is difficult to eliminate them simultaneously.

\subsection{Line Segment Post-processing with 3D Gaussians}
\label{sec:3.3}

To eliminate the aforementioned drawbacks and defects in Sec.~\ref{sec:3.2} while retaining the geometric validity of the reconstructed segments, we propose a targeted post-processing approach using the robust spatial features of 3D Gaussian points. Our method aims to align the segments with the 3D Gaussian distribution that have a strong consistency of spatial features, achieving a more concise and accurate 3D line reconstruction and an abstract representation of the 3D Gaussian distribution. For the convenience of explanation, the \textbf{Gaussians} mentioned in the following text all refer to the Gaussian center points.

\begin{figure}[htbp]
    \centering
    \includegraphics[width=\linewidth]{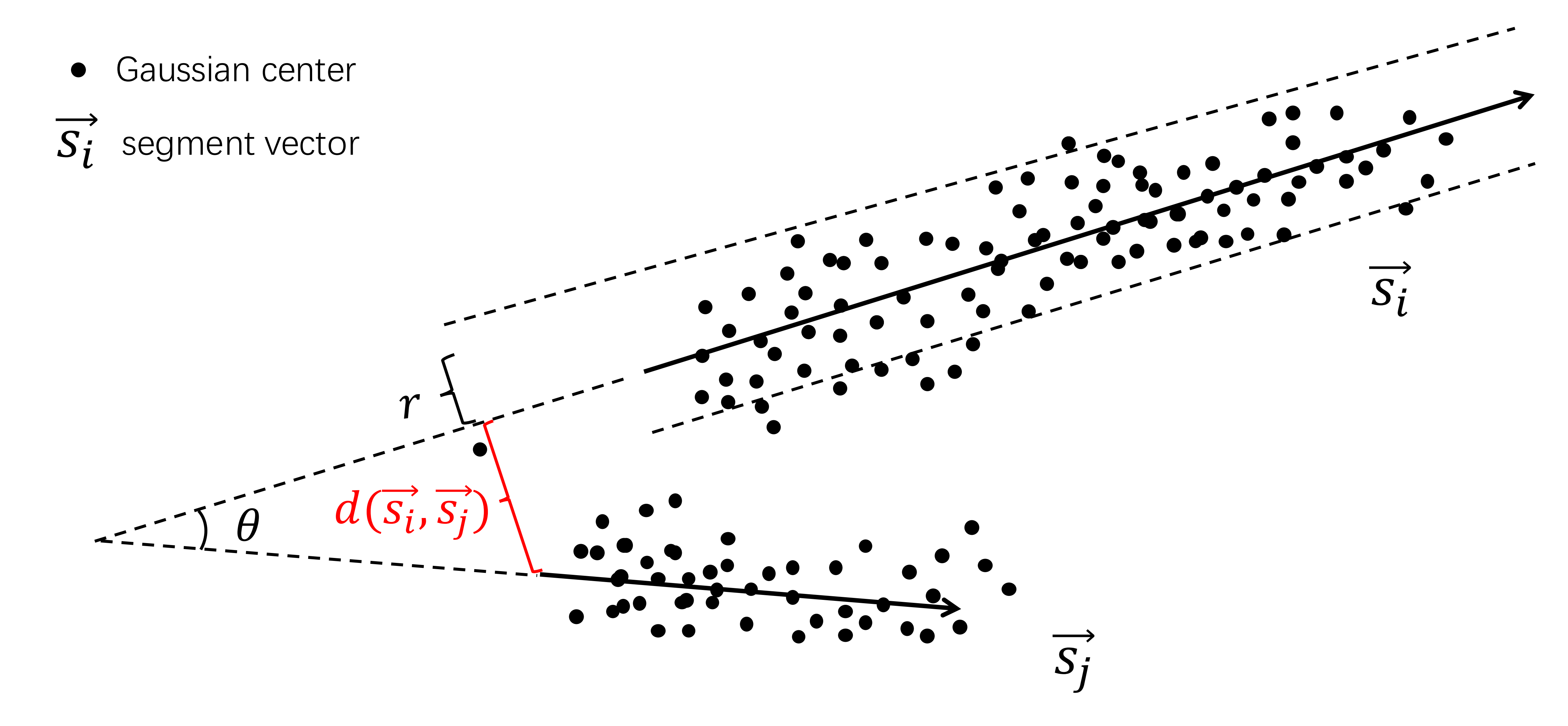}
    \caption{It shows the defination of $C(\Vec{s}, r)$ and the calculation of similarity between two line segments $\Vec{s_i}$ and $\Vec{s_j}$.}
    \label{fig:segment_similarity}
\end{figure}

First in Fig.~\ref{fig:segment_similarity}, we define a cylinder space $C(\Vec{s}, r)$ with the line segment $\Vec{s}$ as its central axis and a radius of $r$. The point set $X(\Vec{s}) = \{x \in C(\Vec{s}, r)\}, X(\Vec{s}) \subseteq G$, where $G$ is the set of all Gaussians center in the scene, represents all Gaussians within the cylinder space. To efficiently obtain each $X(\Vec{s})$, we use an octree structure~\cite{meagher1980octree,meagher1982geometric}. By placing all Gaussians into the octree, we recursively search for nodes whose boundaries intersect with the cylinder, retrieving the Gaussian centers contained within these nodes. This cylindrical space $C$ is used to calculate the number of Gaussians covered by the current line segment. It also allows for the computation of the Gaussian density along the segment, and the RMSE (Root Mean Square Error) of the distances between the covered Gaussians and the line segment which contributes to further processing and evaluating.

For \textbf{position bias} between the line segments and the dense Gaussian regions mentioned in Sec.~\ref{sec:3.2}, we translate the original line segment using the Gaussians within the cylinder through linear regression~\cite{montgomery2021introduction,weisberg2005applied}, considering the efficiency and accuracy. 
\begin{equation}
\label{equ:lr expression}
\mathbf{t} = \arg\min_{\mathbf{t}} \sum_{i=1}^{N} \left\| dist(x', \Vec{s}) - \mathbf{t} \right\|^2
, N = |X(\Vec{s})|\end{equation}
Eq.~\ref{equ:lr expression} shows the formula for linear regression, where $x'$ is the projection of point $x$ onto the orthogonal plane of the line segment $\Vec{s}$. It calculates the Euclidean distance from the projection point to the line segment. $N$ represents the number of Gaussian points covered within the cylindrical space of the current line segment. The final vector $t$ represents the translation vector for the line segment $\Vec{s}$.

\begin{algorithm}[H]
\caption{Binary Search Cropping}
    \begin{algorithmic}[1]
    \label{alg:binary crop}
    \REQUIRE Segment $\Vec{s}$, Octree $\mathbf{o}$, Parameters $\mathbf{p}$
    \STATE idx $\gets 0$
    \STATE start, end $\gets \Vec{s}$
    \STATE end\_density $\gets Density(end, \mathbf{O})$
    \WHILE{True}
        \STATE idx $\gets$ idx + 1
        \STATE mid $\gets$ (start + end) / 2
        \IF{idx $>$ 10 or $\|$start - end$\| < 1^{-4}$}
            \STATE \textbf{break}
        \ENDIF
        \STATE $sub\_seg \gets CreateSegment(mid)$
        \STATE mid\_density $\gets Density(mid, \mathbf{O})$
        \IF{$isDense(mid\_density, end\_density, \mathbf{p})$}
            \STATE end $\gets$ mid
        \ELSE
            \STATE start $\gets$ mid
        \ENDIF
    \ENDWHILE
    \STATE $\Vec{s} \gets$ mid, end
    \end{algorithmic}
\end{algorithm}

\begin{figure}[htbp]
    \centering
    \includegraphics[width=\linewidth]{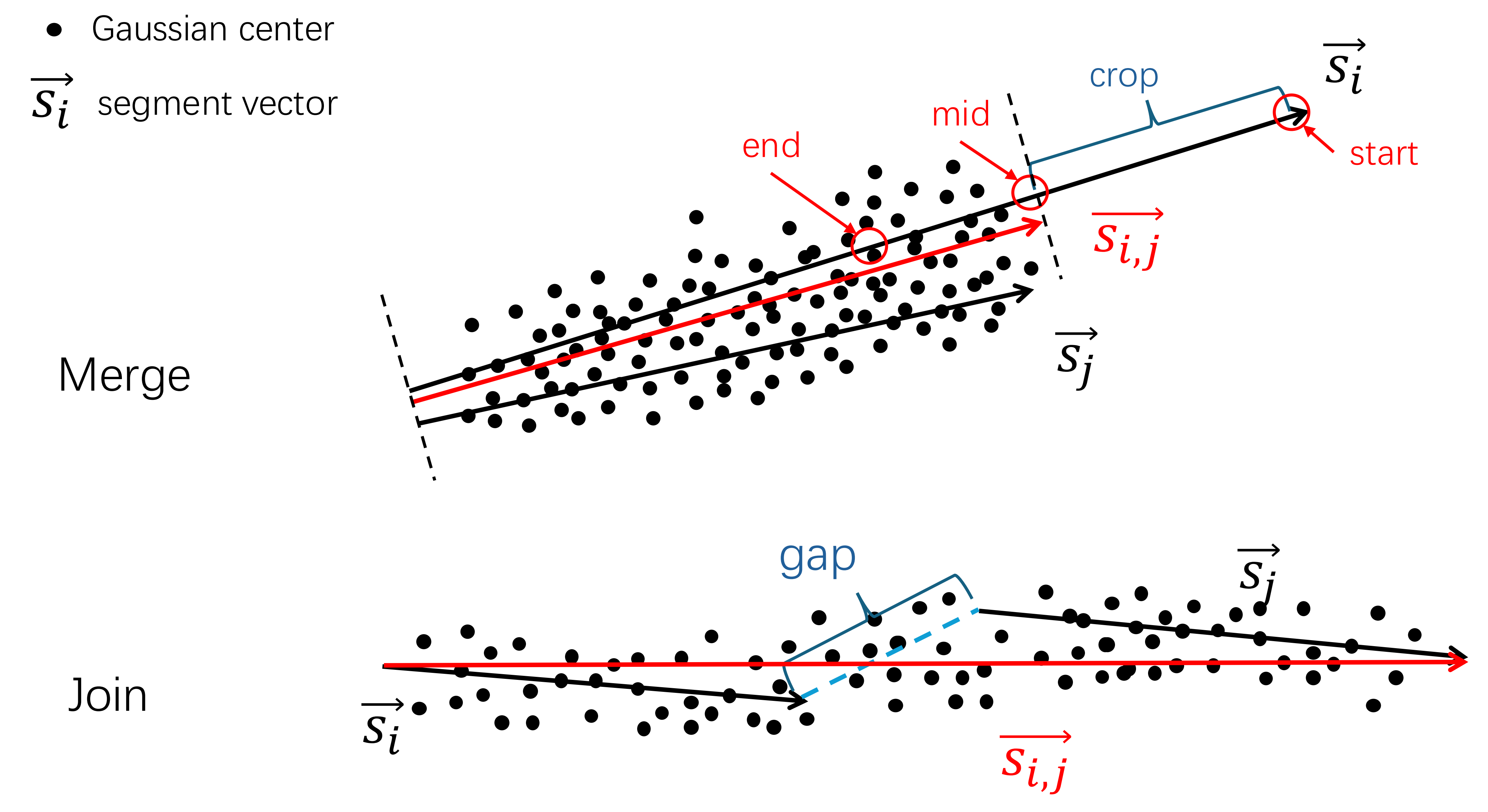}
    \caption{The binary search cropping strategy of single segment $\Vec{s_i}$, and merge-join strategy of two segments $\Vec{s_i}$, $\Vec{s_j}$ within one cluster.}
    \label{fig:segment_cluster}
\end{figure}

To address the \textbf{overextension} issue, we evaluate the endpoints of the line segment and crop the extension. If the Gaussian point density near an endpoint is much lower than the density of the central part of the segment, it suggests that a sub-segment containing this endpoint is the overextension part which should be cropped, so we set that endpoint as the start and the midpoint as the end. Algorithm~\ref{alg:binary crop} shows an implementation of binary search to find the boundary point where the Gaussian density significantly differs on either side, and Fig.~\ref{fig:segment_cluster} shows a visible example of cropping strategy. Binary search is efficient for locating this boundary within limited iterations, as minor length bias are acceptable. The midpoint is chosen as end based on the assumption, validated through experiments, that the overextended section is no more than half the segment’s length.

Since we can calculate the density of each segment, we can identify \textbf{outliers} by setting a global density threshold $\theta$. In Eq.~\ref{equ:density threshold}, we calculate it based on the density $d$ of all segments in the scene:
\begin{equation}
\label{equ:density threshold}
\theta =  \xi \cdot \frac{1}{n} \sum_{i=1}^{n} d_i
\end{equation}
Segments with densities below this threshold are directly removed.

For \textbf{duplication} and \textbf{discontinuity}, we first calculate an affinity matrix (or similarity matrix) between segments. Segments with high similarity are grouped into the same cluster. Within each cluster, segment relationships are then classified as either duplication or discontinuity, and each category is processed accordingly. The mathematical meaning of these kind similarity is as follows: if two segments are very close to each other, they are considered similar and potentially mergeable if they satisfy either of these conditions: 1) When the length ratio of the two segments is large, a small angle between them is less critical; 2) When the segment lengths are similar, the angle between them should be as small as possible. Eq.~\ref{equ:cluster similarity} shows the way to get the similarity of two segment $\Vec{s_i}$ and $\Vec{s_j}$:
\begin{equation}
\label{equ:cluster similarity}
\mathbf{s}\left(\Vec{s_i}, \Vec{s_j}\right) = 
\begin{cases} 
\dfrac{\tanh{(R^2 \cdot \cos \theta)}}{1 + \lambda \cdot {d(\Vec{s_i}, \Vec{s_j})}^2}  & \text{if } \cos \theta < 0.5 \\
0 & \text{otherwise}
\end{cases}
\end{equation}
where $R$ is the length ratio of the longer segment to the shorter one, $\theta$ is the angle between the direction vectors of the two segments, and $d(\Vec{s_i}, \Vec{s_j})$ is the distance from the endpoint of the shorter segment to the line on which the longer segment lies, and weight by $\lambda$. Fig.~\ref{fig:segment_similarity} shows the visual defination of $d(\Vec{s_i}, \Vec{s_j})$.

\begin{algorithm}[H]
\caption{Perform Clustering of 3D Line Segments}
    \begin{algorithmic}[1]
    \label{alg:cluster}
    \REQUIRE Segments $\mathbf{s} = \{\Vec{s_0},\Vec{s_1},\dots\Vec{s_n}\}$, Parameters $\mathbf{p}$
    \ENSURE Clustered Universe $\mathbf{u}$
    
    \STATE $\text{num} \gets |\mathbf{s}|$
    \STATE $\text{edges} \gets \text{similarity between every two segments}$
    
    \STATE Sort $\text{edges}$ by similarity in ascending order
    \STATE $\mathbf{u} \gets \text{CLUniverse}(\text{num})$
    \STATE $\text{threshold} \gets [\mathbf{p}.\text{cluster\_c}] \times \text{num}$
    
    \FOR{each $\text{edge} \in \text{edges}$}
        \STATE $a \gets \mathbf{u}.\text{find}(\text{edge}.\Vec{s_i})$
        \STATE $b \gets \mathbf{u}.\text{find}(\text{edge}.\Vec{s_j})$
        \IF{$a \neq b$ and $\text{edge.w} \leq \text{threshold}[a]$ and $\text{edge.w} \leq \text{threshold}[b]$}
            \STATE $\mathbf{u}.\text{join}(a, b)$
            \STATE $r \gets \mathbf{u}.\text{find}(a)$
            \STATE $\text{threshold}[r] \gets \text{edge.w} + \frac{\mathbf{p}.\text{cluster\_c}}{\mathbf{u}.\text{size}(r)}$
        \ENDIF
    \ENDFOR
    
    \STATE \textbf{return} $\mathbf{u}$
    \end{algorithmic}
\end{algorithm}
After getting the similarity of each two segments, we apply the clustering Algorithm~\ref{alg:cluster}, inspired from work~\cite{hofer2017efficient}. This clustering method is similar to a union-find structure, treating each segment as a node and the similarity between two segments as an edge weight. First, all edges are sorted by similarity, and segments connected by high-similarity edges are sequentially grouped into the same cluster. The similarity threshold for adding to each cluster is updated to control the number of segments included. In the end, each segment is assigned to a cluster, where all segments have high similarity mentioned in Eq.~\ref{equ:cluster similarity}.

As Fig.~\ref{fig:segment_cluster} shows, for a pair of segments within a cluster, we have two kinds of option. 1) if they have overlapping parts, a merge operation is attempted. If the region between the segments contains a sufficient number of Gaussian centers, we try shifting the longer segment toward the shorter one proportionally to their lengths and compare the result with the original long segment, retaining the one with the smallest $R = \dfrac{E_{rms}}{N}$, where $E_{rms}$ is the RMSE value of the segment, $N$ is the number of Gaussians that segment covers. 2) If the two segments do not overlap, a join operation is attempted. If the gap segment between them has sufficient point density and creating a new segment connecting their farthest endpoints would improve the evaluation result, we retain the newly created segment. Eq.~\ref{equ:overlap} shows the logic for determining whether there is overlap:
\begin{equation}
\label{equ:overlap}
(\mathbf{p} - P_1) \cdot \Vec{d} > 0 \cup (\mathbf{p} - P_2) \cdot \Vec{d} < 0
\end{equation}
where $P_1$ and $P_2$ are two endpoints of longer segment, $\Vec{d}$ is the direction of longer segment, $\mathbf{p}$ is one of the endpoints of shorter segment.

\begin{figure*}[htbp]
    \centering
    \includegraphics[width=\linewidth]{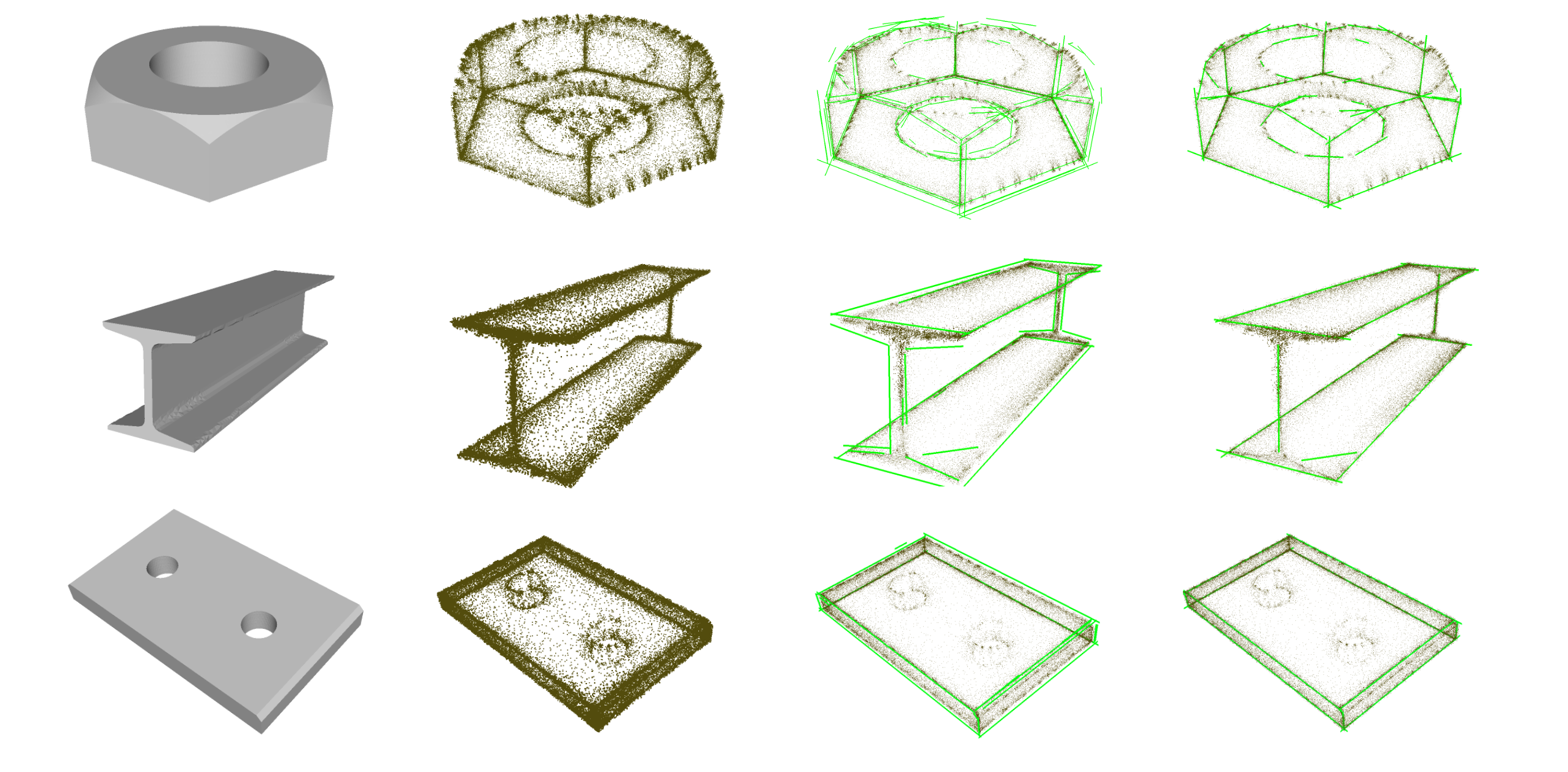}
    \includegraphics[width=\linewidth]{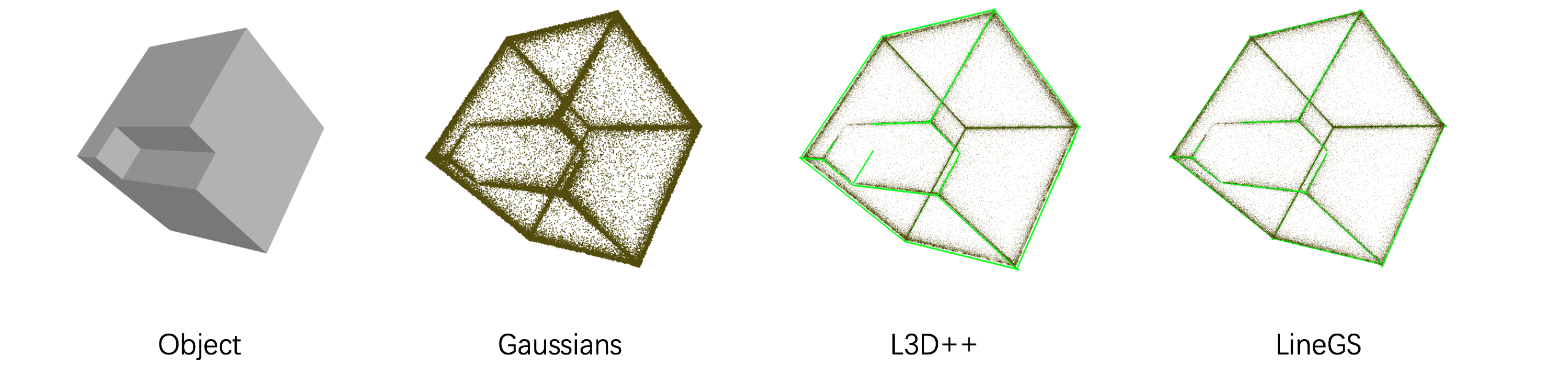}
    \caption{\textbf{Qualitative results on ABC-NEF~\cite{ye2023nef}.} The proposed method accurately addresses issues, such as redundant segments and misalignment with the actual boundaries of Gaussians. While it may sacrifice some boundary details, this trade-off is acceptable for the Gaussians.}
    \label{fig:abc}
\end{figure*}

\section{Evaluation}

In this section, we introduce our approach for quantitatively evaluating whether 3D line segments effectively represent the spatial and color features of the scene, as well as the distribution of 3D Gaussian model centers. We propose a custom evaluation method, integrating insights from previous 3D line reconstruction work~\cite{hofer2017efficient, wei2022elsr} and assessing their performance on this task. Both visualizations and quantitative metrics demonstrate the effectiveness of our approach.

\subsection{Experiment Setup}
\textbf{Datasets.} 
First, for simple scenes and models, we used the ABC-NEF~\cite{ye2023nef} dataset, a subset of the ABC~\cite{koch2019abc} dataset commonly used in 3D line or boundary tasks. This dataset includes more than 100 CAD models, along with camera calibration parameters and images formatted for NeRF~\cite{mildenhall2021nerf}. Since the Gaussian splatting model also supports NeRF input, we can train on these CAD models to obtain Gaussian models for evaluation. For more complex actual scenes, including indoor and outdoor settings, we used datasets from the original Gaussian model work~\cite{kerbl3Dgaussians}, which includes the indoor scenes \textit{Playroom}, \textit{Drjohnson} from  Deep Blending dataset~\cite{hedman2018deep}, \textit{Room}, \textit{Counter} from the Mip-NeRF360 dataset~\cite{barron2022mip}, the outdoor scenes \textit{Train}, \textit{Truck} from Tanks\&Temples~\cite{knapitsch2017tanks} and \textit{Herz-Jesu-25} scene from DenseMVS~\cite{strecha2008benchmarking}.

\textbf{Baselines.} 
Our proposed approach utilizes a pre-trained Gaussian model to post-process 3D line segments reconstructed by previous geometry-based methods. To demonstrate the effectiveness of this post-processing, we should generate both the Gaussian model and initial 3D segments from the same input or original data samples, specifically, 2D images and corresponding SfM result, then apply post-processing and compare the results with the initial segments to validate the method's effectiveness.
We use the default training parameters and method from original work~\cite{kerbl3Dgaussians} to train the Gaussian models. For the initial 3D line segment reconstruction, we choose L3D++~\cite{hofer2017efficient} and ELSR~\cite{wei2022elsr} as a representative, two state-of-the-art methods capable of efficiently reconstructing 3D line segments directly from SfM and images, using the default line detector LSD~\cite{ipol.2012.gjmr-lsd}.

\begin{table*}[htbp]
    \renewcommand\arraystretch{1.3}
    \caption{\textbf{Quantitive evaluation with L3D++~\cite{hofer2017efficient}.} Overall, the proposed method improves the representational ability of 3D line segments on Gaussians. The evaluation is under  cylinder space with 10cm radius.}
    \label{tab:abc}
    \begin{center}
        \begin{tabular}{|c|cccc|cccc|c|}
        \toprule 
        \multirow{2}{*}{Dataset}       & \multicolumn{4}{c|}{L3D++}                                        & \multicolumn{4}{c|}{Ours + L3D++}  & \multirow{2}{*}{Improvement}  \\
                             & $E_{rms} \downarrow$ & $R_{covered} \uparrow$ & $R_L \downarrow$ & $Score \uparrow$                  & $E_{rms} \downarrow$ & $R_{covered} \uparrow$ & $R_L \downarrow$ & $Score \uparrow$ & \\ \hline
        ABC-NEF~\cite{ye2023nef}     & 4.72                 & \textbf{92.0}        & 1.06            
           & 7.784                 & \textbf{4.21}        & 90.1                 & \textbf{0.87}  
           &\textbf{9.431}        & 21.2\% \\ \hline
        playroom~\cite{hedman2018deep}& 5.73                 & \textbf{65.6}                & 79.44       & 7.847                 & \textbf{5.26}        & 60.1                 & \textbf{47.94}       & \textbf{8.421}        & 7.3\% \\ \hline
        drjohnson~\cite{hedman2018deep}& 5.62                 & \textbf{60.3}                & 60.34       & 7.752                 & \textbf{5.44}        & 55.6                 & \textbf{31.98}       & \textbf{8.530}        & 10.1\% \\ \hline
        counter~\cite{barron2022mip}& 5.54                 & \textbf{45.2}                & 56.20       & 5.944                 & \textbf{5.21}        & 38.01                 & \textbf{10.58}       & \textbf{8.497}        & 42.9\% \\ \hline
        room~\cite{barron2022mip}& 5.61                 & \textbf{43.5}                & 57.35       & 5.663                 & \textbf{5.02}        & 37.5                 & \textbf{21.31}       & \textbf{6.730}        & 18.6\% \\ \hline
        train~\cite{knapitsch2017tanks}& 7.51                 & \textbf{24.5}                & 25.18       & 3.502                 & \textbf{6.53}        & 18.1                 & \textbf{5.77}        & \textbf{4.697}        & 34.1\% \\ \hline
        truck~\cite{knapitsch2017tanks}& 5.46                 & \textbf{23.4}                & 36.38       & 3.458                 & \textbf{4.99}        & 20.9                 & \textbf{15.05}       & \textbf{4.208}          & 21.7\% \\  
        \bottomrule
        \end{tabular}
    \end{center}
\end{table*}
\textbf{Metrics.}
Evaluating the quality of scene representation is typically done by calculating the error with respect to ground-truth data. However, aside from CAD models like those in the ABC-NEF~\cite{ye2023nef} dataset, ground-truth values are challenging to obtain for real-world scenes. In contrast, evaluating the representation of Gaussian center distributions is more feasible, as we have a high-precision Gaussian model trained for this purpose. We use multiple metrics to assess the quality of representation.
\begin{equation}
\label{equ:score}
\text{score} =  \dfrac{\lambda \cdot R_{covered}}{\log(1 + E_{rms}) \cdot \log(1 + \mathrm{R_L})}
\end{equation}
We aim for line segments that reflect Gaussian points with similar features, clustering near the segment. Since our post-processing method does not significantly alter the segment's position or direction, and the initial 3D segments generated geometrically are deemed reliable, an ideal 3D line segment should: 1) have Gaussian points densely clustered along it, measured by $E_{rms}$, with units in centimeters, and 2) exhibit a Gaussian point coverage percentage $R_{covered}$ proportional to its length. For segments with similar lengths and quantities, higher Gaussian point coverage is preferable, so we measure this by the ratio of Gaussian coverage percentage $R_{covered}$ to length metric $\mathrm{R_L}$.Due to varying scene sizes, segment lengths are not uniformly scaled, so we use the ratio of the actual segment length to the logarithm of the Gaussian point count as the $\mathrm{R_L}$ metric above, Eq.~\ref{equ:score1} shows how to get this metric, where $l$ is the length of segment and $X(\Vec{s})$ is the point set of cylinder space of the segment:
\begin{equation}
\label{equ:score1}
\mathrm{R_L} = \dfrac{\sum_{i=1}^{n} l_i}{\log\left(\sum_{i=1}^{n} |X(\Vec{s_i})|\right)}
\end{equation}
To quantify this, we use the score in Eq.~\ref{equ:score}, where $\lambda$ is a scaling factor.

\textbf{Implementation Details.}
Since the ABC-NEF~\cite{ye2023nef} dataset provides standard camera parameters, we substitute downsampled Gaussian centers as sparse point clouds, allowing us to use these as inputs to the 3D line reconstruction methods without relying on SfM~\cite{schoenberger2016sfm} to generate camera parameters and sparse point clouds. The weight of similarity function in Eq.~\ref{equ:cluster similarity} is $\lambda = 2 / r^2$, where radius $r$ of cylinder space of segments is 3cm for ABC-NEF~\cite{ye2023nef}, and 5cm for other datasets. The scaler in Eq.~\ref{equ:density threshold} is $\xi = 0.02$. And scaler in Eq.~\ref{equ:score} is $\lambda = 0.1$ for ABC-NEF~\cite{ye2023nef}, and $\lambda = 1$ for other datasets. To enhance algorithm efficiency and evaluation effectiveness, the height of the octree is set to 10, and its bounding box size is based on the endpoint range of the initial 3D segments. Gaussian centers outside the bounding box are excluded and do not participate in the evaluation.

\subsection{Results}

\begin{figure}[htbp]
    \centering
    \includegraphics[width=\linewidth]{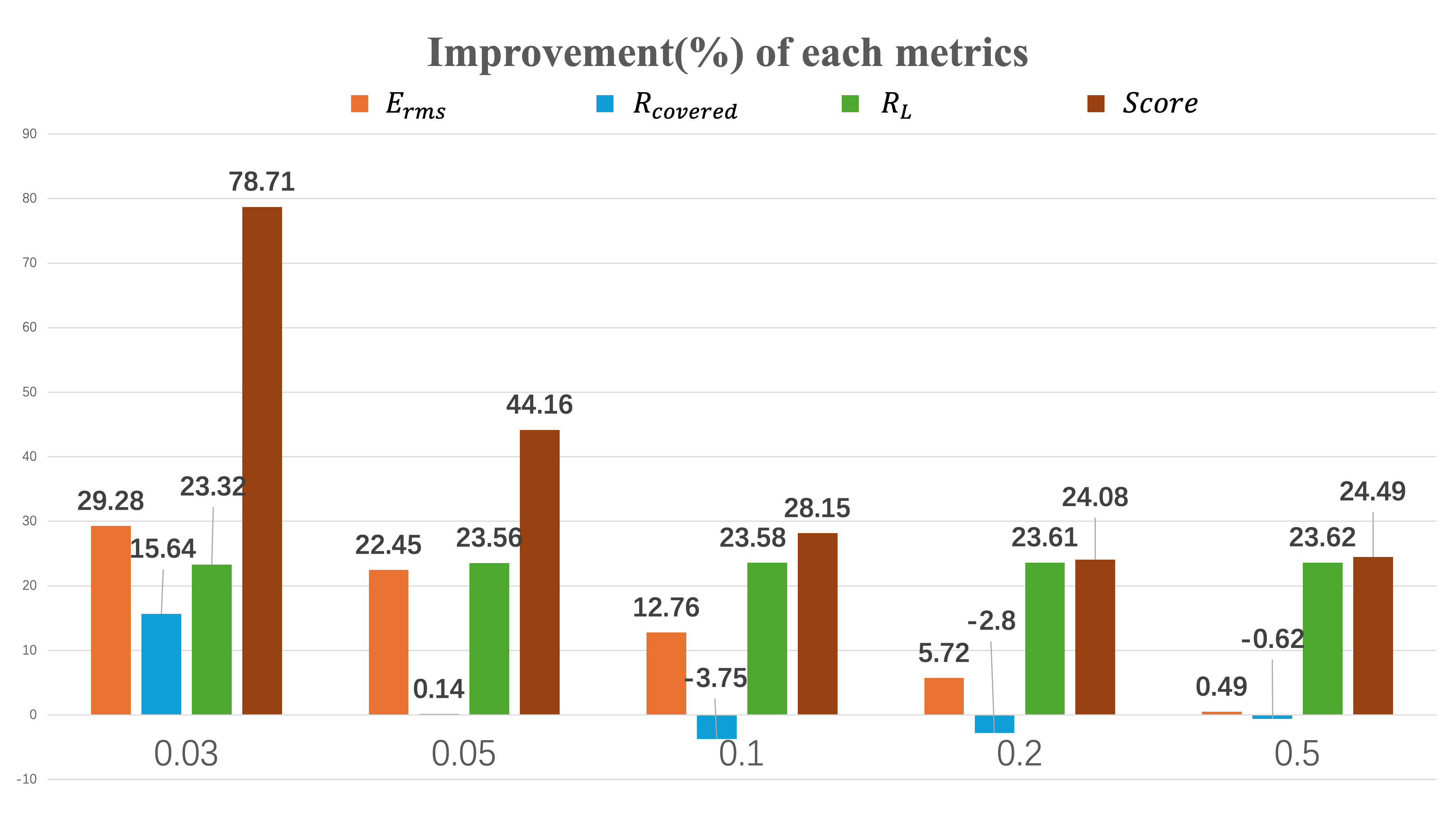}
    \caption{\textbf{Evaluation results with different radii.} Here, we present evaluations using cylinders of varying radii to assess the line segments. The results show noticeable differences across radii, corresponding to the explanation provided in the main text. The x-axis represents the values of radii, measured in meters, and the y-axis shows the values of each parameter. }
    \label{fig:abc_eval}
\end{figure}

\textbf{Evaluation on ABC-NEF~\cite{ye2023nef}.} The quantitative results are reported in Tab.~\ref{tab:abc} and some sample qualitative results in Fig.~\ref{fig:abc}. Overall, our proposed method improves the fit of 3D line segments to Gaussian centers, resulting in better segment representation. Across the entire ABC-NEF dataset of 115 models, the total $E_{rms}$ decreased by 10.8\%, and the Gaussians fitting score increased by 21.2\%. The decrease $E_{rms}$ in for Gaussian centers near line segments indicates that the segments are more accurately aligned with the Gaussian center distribution. Additionally, the shorter segment lengths suggest a better fit to the boundaries of the Gaussian distribution, achieving a more concise abstract representation. Compared to the L3D++~\cite{hofer2017efficient}, our post-processed results cover fewer Gaussian centers, which aligns more closely with our goals. We attribute this difference to two main factors. 1) L3D++~\cite{hofer2017efficient} produces more segments, but some are unsuitable or inaccurate in the context of the Gaussian model, and the points represented by these segments may lack spatial consistency in Gaussian centers. By contrast, the segments retained or modified through our method represent Gaussian centers with greater spatial and feature consistency, benefiting from the geometry-guided line reconstruction algorithms, among them, some Gaussian points that were originally represented will inevitably be discarded. 2) This result is also influenced by the radius $r$ of the cylinder space used during evaluation. 
\begin{figure}[htbp]
    \centering
    \includegraphics[width=\linewidth]{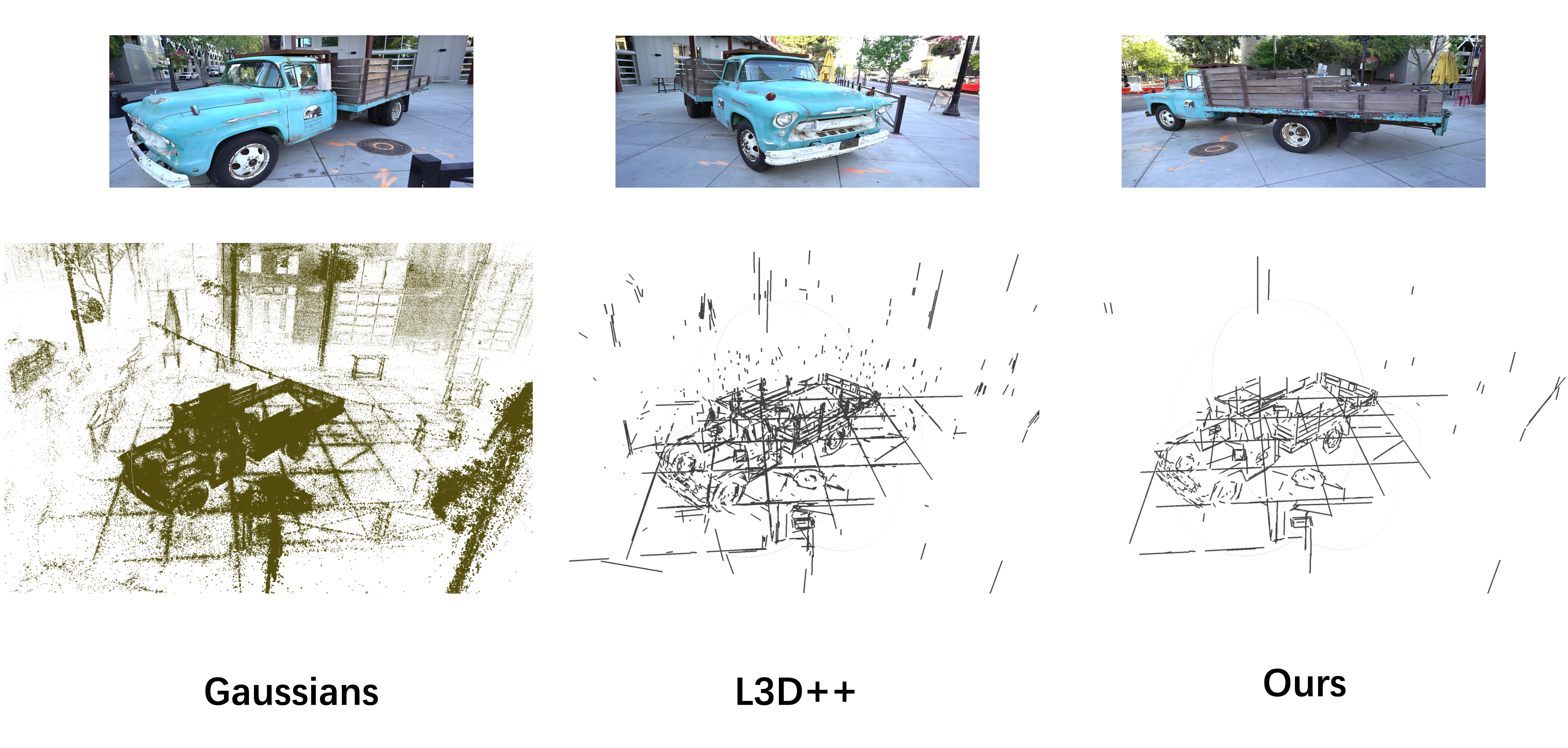}
    \caption{\textbf{Evaluation results on truck.} As we can see, segments produced by our method are clearer, more concise, and align well with the Gaussian model's distribution, retaining only regions with a high density of Gaussian points.}
    \label{fig:truck}
\end{figure}
With a smaller $r$, adjustments made during processing yield a greater improvement over the initial segments. With a larger $r$, adjustments may yield minimal improvement in point coverage or even result in a decline, as our adjustments are limited to a smaller region. Fig.~\ref{fig:abc_eval} shows the difference results corresponding to different $r$ of the cylinder space in evaluation.

\textbf{Evaluation on Indoor/Outdoor Scene.} Our proposed method demonstrates significant improvements in both indoor and outdoor scenes, with quantitative results shown in Tab.~\ref{tab:abc} and qualitative results for truck~\cite{knapitsch2017tanks} displayed in Fig.~\ref{fig:truck}. Overall, our method consistently enhances the representation of 3D Gaussians in real-world scenes. The trend in evaluation parameters mirrors that observed in the ABC-NEF~\cite{ye2023nef}: there is some loss in point coverage $R_{covered}$, but the spatial consistency in representing Gaussians surpasses that of the unprocessed segments, with score improvements ranging from 7\% to 43\%. 

\begin{table}[htbp]
    \renewcommand\arraystretch{1.3}
    \caption{\textbf{Quantitive evaluation on Herz-Jesu-25~\cite{strecha2008benchmarking}.} For both geometry-guided methods, our proposed method improves the representational ability. The evaluation is under  cylinder space with 10cm radius.}
    \label{tab:herz}
    \begin{center}
        \begin{tabular}{c|cccc}
        \toprule 
        Method       & $E_{rms} \downarrow$ & $R_{covered} \uparrow$ & $R_L \downarrow$ & $Score \uparrow$       \\ \hline
        L3D++~\cite{hofer2017efficient}& 5.55          & \textbf{39.6} & 10.76          &    8.555           \\
        Ours + L3D++            & \textbf{5.43} & 37.6          & \textbf{6.33}  & \textbf{10.142}                               \\ 
        \hline
        ELSR~\cite{wei2022elsr}& 5.67          & \textbf{45.8} & 13.98          & 8.916          
                \\
        Ours + ELSR             & \textbf{5.55} & 43.2          & \textbf{8.57}  & \textbf{10.186}                               \\
        \bottomrule
        \end{tabular}
    \end{center}
\end{table}

\begin{figure}[htbp]
    \centering
    \includegraphics[width=\linewidth]{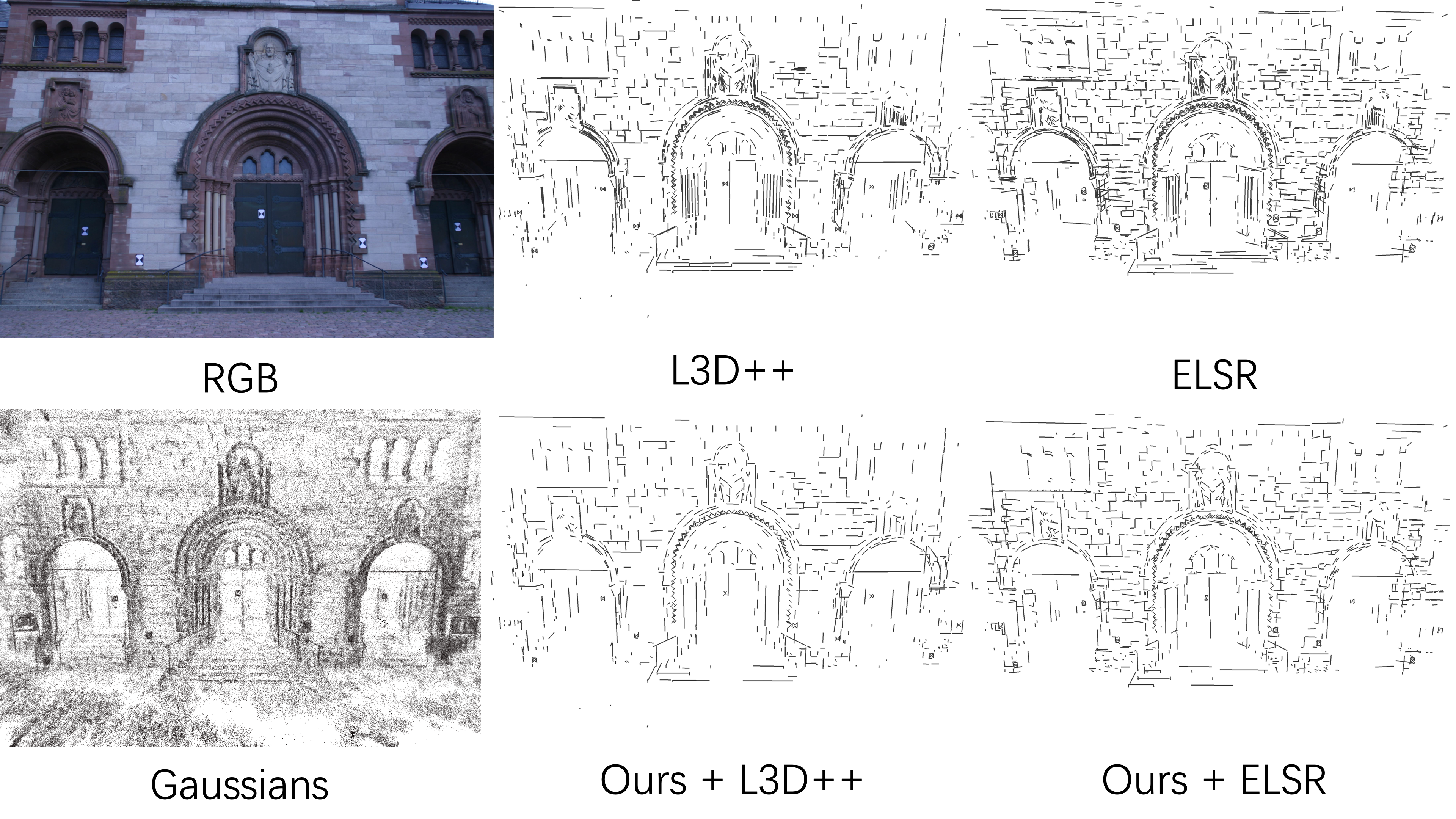}
    \caption{\textbf{Qualitative results on Herz-Jesu-25~\cite{strecha2008benchmarking}.} We can see that our method preserves complete and accurate spatial information while retaining line segments that represent areas with sufficient Gaussian density. Additionally, it optimizes the position of the segments, resulting in a more refined abstract representation of the Gaussians.}
    \label{fig:herz}
\end{figure}
To validate the generalizability of our method, we also reproduced the ELSR~\cite{wei2022elsr} method. Compared to L3D++~\cite{hofer2017efficient}, ELSR~\cite{wei2022elsr} generates more initial line segments with smaller spatial position errors relative to ground truth. We compared the results of both methods on the Herz-Jesu-25~\cite{strecha2008benchmarking} scene, as shown in Tab.~\ref{tab:herz}, each processed with our approach. The results demonstrate significant improvements across different initial line segments, with a 18.6\% improvement for L3D++~\cite{hofer2017efficient} segments and a 14.2\% improvement for ELSR~\cite{wei2022elsr} segments. Fig.~\ref{fig:herz} shows the initialization methods of the two different line segment methods and their corresponding post-processed results.
\section{Conclusion}

In this work, we propose a 3D line reconstruction method that better represents the distribution of Gaussian model centers. Our approach is based on existing geometry-guided 3D line reconstruction and incorporates post-processing with the Gaussian model, making it straightforward to implement. Experiments demonstrate that this method optimizes initial line segments and enhances the fit to the Gaussian center distribution, providing a meaningful abstraction and representation of the Gaussian model. While the improvements from post-processing may be influenced by more advanced line reconstruction methods, incorporating the unique non-spatial features of Gaussian models in the future could further enhance this approach. \\
\textbf{Acknowledgements.} This work was supported by the School of Computing at NUS.

\end{document}